\documentclass[letterpaper, 10 pt, conference]{ieeeconf}

\IEEEoverridecommandlockouts

\usepackage{algorithmic}
\usepackage{algorithm}
\usepackage{array}

\usepackage{textcomp}
\usepackage{stfloats}
\usepackage{subfigure}

\usepackage{url}
\usepackage{verbatim}
\usepackage{graphicx}
\usepackage{cite}
\usepackage{hyperref}
\usepackage[hyphenbreaks]{breakurl}
\usepackage{amsfonts, amssymb}
\usepackage{mathtools}
\usepackage{booktabs}
\usepackage{multirow}
\usepackage{multicol}
\usepackage[table]{xcolor}

\title{\LARGE \bf
VLM-MPPI: Grounding Natural Language in Behaviorally Diverse Trajectories for Aerial Navigation
}

\author{Hanbing Zhang$^{1}$, Fangguo Zhao$^{1}$, Zerui Li$^{2}$, Xin Guan$^{1}$, Peng Cheng$^{1}$, Shuo Li$^{1}$\thanks{$^{1}$Authors are with the College of Control Science and Engineering, Zhejiang University, Hangzhou 310027, China
        {\tt\small shuo.li@zju.edu.cn}}\thanks{$^{2}$Zerui Li is with Australian Institute for Machine Learning, Adelaide University.}\thanks{This work was supported in part by the National Natural Science Foundation of China (NSFC) under Grants W2511069 and 62088101, and in part by the Zhejiang Key Laboratory of Airspace Perception and Autonomous Unmanned Systems.}}

\begin{document}

\maketitle
\thispagestyle{empty}
\pagestyle{empty}

\begin{abstract}

We present a hierarchical UAV navigation framework that aligns natural-language intent with dynamically feasible flight behaviors in cluttered indoor environments. To bridge the gap between abstract semantics and low-level control, we employ a parallelized ensemble of six behavior-conditioned Model Predictive Path Integral (MPPI) planners. Crucially, by designing mode-specific guiding costs and sampling biases, we induce distinct trajectory modes that converge to unique behavioral means, yielding a compact set of intentionally diverse candidates rather than mere stochastic variations. We project these 3D candidates onto the onboard first-person-view RGB stream, turning language grounding into a visual action selection problem. A pretrained vision--language model (VLM) asynchronously selects the candidate index given the overlaid FPV image and a natural-language prompt, while MPPI replans at \textbf{20\,Hz} and a PID-based low-level controller tracks the selected trajectory.
We implement the full pipeline in NVIDIA Isaac Sim and on a real-world quadrotor platform equipped with LiDAR and RGB sensing. Experiments in both simulation and real-world flights show semantically meaningful behavior diversity, robust language alignment despite VLM latency, and safe, repeatable flight across all modes, achieving \textbf{100\%} task success in our evaluated scenarios.
\end{abstract}

\section{Introduction}

In recent years, unmanned aerial vehicles (UAVs) have evolved from remotely piloted platforms into autonomous agents operating in complex, cluttered environments\cite{zhou2020ego}. As they are increasingly used for tasks such as indoor inspection and search-and-rescue, human--robot interaction is shifting from purely geometric specifications toward rich, high-level semantic instructions\cite{driess2023palm,gu2022vision}. Beyond providing waypoints or goal poses, users now expect to influence how the vehicle moves using natural language commands such as ``follow the corridor quietly,'' ``inspect the object on the left,'' or ``move conservatively through the crowd.'' Executing such commands requires translating abstract human intent into precise, dynamically feasible control actions, a challenge that is particularly acute in constrained indoor environments\cite{duan2022survey}, where the UAV must interpret visual context while strictly respecting kinematic and dynamic limits to ensure safe operation.

\begin{figure}[t]
    \centering
    \includegraphics[width=0.95\linewidth]{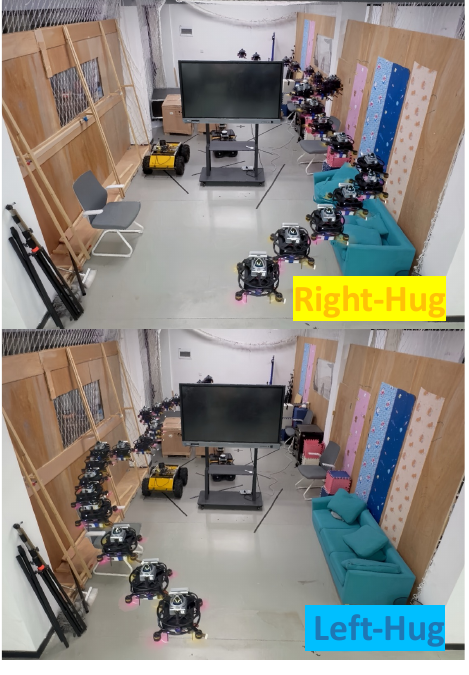}
    \vspace{-0.2cm}
    \caption{\textbf{Real-world Deployment.} Composite flight trajectories in a $7\,\text{m} \times 4\,\text{m} \times 3\,\text{m}$ cluttered environment. The system executes topologically distinct behavioral modes (\textbf{Right-Hug} and \textbf{Left-Hug}) tailored to varying prompts, demonstrating robust alignment between semantic intent and feasible motion.}
    \label{fig:real_world_teaser}
    \vspace{-1.5em}
\end{figure}

Over the past decade, indoor UAV navigation has evolved from geometric planning to kinodynamic trajectory generation with frequent replanning. Minimum-snap trajectories~\cite{mellinger2011minimum} enable real-time, dynamically feasible quadrotor motion through waypoints under corridor and input constraints. Fast-Planner~\cite{zhou2019robust} combines kinodynamic search with efficient B-spline optimization and time adjustment for fast, non-conservative replanning, while EGO-Planner~\cite{zhou2020ego} improves efficiency via an ESDF-free gradient-based formulation guided by a collision-free reference and lightweight obstacle extraction. In parallel, MPPI~\cite{williams2017model} optimizes controls through massively parallel rollouts, and prior studies have shown performance comparable to optimization-based MPC in challenging setting\cite{zhao2025rethinking}; PA-MPPI~\cite{Pa-mppi} adds perception-aware costs to reveal unknown space when the goal is occluded, and AERO-MPPI~\cite{AERO-MPPI} uses LiDAR-derived anchors to guide parallel MPPI optimizers toward diverse homotopy classes. Despite these advances, mainstream navigation pipelines remain largely semantics-agnostic, with behavior dictated by hand-designed objectives and manually crafted cost terms. As a result, natural language cannot directly specify flight style without manual cost design and tuning.

To overcome the semantic limitations of traditional navigation frameworks, Vision-Language Models (VLMs) such as GPT-4o~\cite{achiam2023gpt} and Gemini 2.5~\cite{comanici2025gemini} have recently been integrated into autonomous systems. Pioneering works such as LM-Nav~\cite{Lm-nav} and CLIP-Nav~\cite{Clip-nav} leverage pre-trained models to achieve zero-shot landmark navigation without task-specific training. Subsequent systems move toward context-aware planning; for example, CoNVOI~\cite{Convoi} and Language-as-Cost~\cite{language} translate visual semantics into risk-aware, static costmaps that guide local planners. In aerial and large-scale navigation, foundation models are predominantly employed as high-level global planners. Specifically, architectures like Online-VLM-Planner~\cite{garg2026online} and UAV-VLPA*~\cite{UAV-VLPA} generate discrete waypoints, whereas others are combined with low-level tracking controllers for tasks such as autonomous driving~\cite{visiopath,ComposableNav}. However, while effectively harnessing semantic priors, these methods predominantly yield geometric outputs, such as discrete waypoints~\cite{rana2023sayplan} or static costmaps~\cite{chen2023open, huang2023voxposer}, which is derived from simplified kinematic models. Consequently, they fail to explicitly enforce the continuous dynamic feasibility and inertial constraints essential for safe aerial navigation. Moreover, the inherent inference latency of large foundation models fundamentally precludes their use in the high-frequency feedback loops essential for flight stabilization.

To bridge this gap, we introduce a hierarchical framework that distinctly separates high-latency semantic reasoning from high-frequency dynamic generation. Instead of prompting a model to synthesize geometric paths, we employ a parallel MPPI stack to continuously generate a compact set of behaviorally diverse, dynamically feasible candidates. By projecting these 3D trajectories onto the onboard FPV images, we let the VLM select the optimal trajectory that  guides the UAV toward the target while obeying the segmentation instructions.The key contributions of our work are as follows:

\begin{itemize}
    \item \textbf{Hierarchical language-conditioned VLM--MPPI navigation framework.} We reformulate language-conditioned UAV navigation as a trajectory selection problem, in which a Vision–Language Model selects the optimal trajectory from dynamically feasible candidates generated by MPPI to guide the UAV toward the target while complying with semantic instructions.

    \item \textbf{Behaviorally Diverse MPPI Ensemble for Trajectory Generation.} We introduce a behaviorally diverse MPPI ensemble that exposes a discrete set of dynamically feasible motion primitives, enabling language-conditioned trajectory selection by a Vision–Language Model.

    \item \textbf{Simulation and Real-World Validation.} We implement the proposed VLM--MPPI stack in NVIDIA Isaac Sim and on a real-world quadrotor platform equipped with LiDAR and RGB sensing. Extensive experiments in both simulation and hardware demonstrate semantically meaningful behavioral diversity across modes, robust language alignment despite VLM latency, and safe, repeatable flight in cluttered indoor scenes.

\end{itemize}

\section{Preliminaries}
\label{sec:preliminaries}

In this section, we briefly review the standard Model Predictive Path Integral (MPPI) algorithm~\cite{williams2017model}, which serves as the mathematical foundation for our trajectory generators. Unlike traditional gradient-based Model Predictive Control (MPC) that often struggles with non-convex spatial landscapes and requires strictly differentiable cost functions, MPPI leverages parallel Monte Carlo sampling to naturally handle complex environments and non-differentiable constraints.

We consider a generic discrete-time nonlinear dynamical system:
\begin{equation}
    \mathbf{x}_{t+1} = \mathcal{F}(\mathbf{x}_t, \mathbf{u}_t),
    \label{eq:generic_dyn}
\end{equation}
where $\mathbf{x}_t \in \mathbb{R}^n$ is the system state and $\mathbf{u}_t \in \mathbb{R}^m$ is the control input. In the context of our aerial navigation task, we define the state as $\mathbf{x}_t = [\mathbf{p}_t^\top, \mathbf{v}_t^\top, \psi_t]^\top \in \mathbb{R}^7$, comprising the 3D position, linear velocity, and yaw angle. The control input is defined as $\mathbf{u}_t = [\mathbf{a}_t^\top, \dot{\psi}_t]^\top \in \mathbb{R}^4$, representing the command linear acceleration and yaw rate. Derived from the information-theoretic formulation of stochastic optimal control, MPPI seeks to find a nominal control sequence $\mathbf{U} = \{\mathbf{u}_0, \dots, \mathbf{u}_{H-1}\}$ over a finite horizon $H$ that minimizes the expected cost:
\begin{equation}
    J(\mathbf{U}) = \mathbb{E}_{\mathbb{Q}} \left[ \phi(\mathbf{x}_H) + \sum_{t=0}^{H-1} \left( q(\mathbf{x}_t) + \frac{1}{2} \mathbf{u}_t^\top \boldsymbol{\Sigma}^{-1} \mathbf{u}_t \right) \right].
    \label{eq:sec}
\end{equation}
where $\phi(\mathbf{x}_H)$ is the terminal state penalty, $q(\mathbf{x}_t)$ is the state-dependent running cost, and the quadratic term penalizes control effort scaled by a positive definite covariance matrix $\boldsymbol{\Sigma}$. The expectation $\mathbb{E}_{\mathbb{Q}}[\cdot]$ is taken over the trajectory distribution $\mathbb{Q}$, induced by injecting Gaussian noise into the nominal sequence $\mathbf{U}$ and propagating the dynamics in~\eqref{eq:generic_dyn}.

In practice, MPPI approximates this expectation by simulating $N$ parallel noisy trajectories. At each time step, control perturbations $\boldsymbol{\epsilon}^{(n)}_t \sim \mathcal{N}(\mathbf{0}, \boldsymbol{\Sigma})$ are sampled to generate $N$ candidate control sequences $\mathbf{u}^{(n)}_t = \mathbf{u}_t + \boldsymbol{\epsilon}^{(n)}_t$. The dynamics are rolled out for each sequence to compute the cumulative trajectory cost $S^{(n)}$. The nominal control sequence is then updated via a probability-weighted moving average:
\begin{equation}
    \mathbf{u}_t \leftarrow \mathbf{u}_t + \eta \frac{\sum_{n=1}^{N} w^{(n)} \boldsymbol{\epsilon}^{(n)}_t}{\sum_{n=1}^{N} w^{(n)}},
    \label{eq:mppi_update}
\end{equation}
where $\eta \in (0, 1]$ dictates the update step size, and the importance weights $w^{(n)}$ follow a Boltzmann distribution:
\begin{equation}
    w^{(n)} = \exp\left(-\frac{1}{\lambda}\left(S^{(n)} - \min_{j} S^{(j)}\right)\right).
    \label{eq:forth}
\end{equation}
where $\lambda > 0$ strictly regulates the exploration-exploitation trade-off.
While this standard MPPI formulation is highly effective for single-objective optimization, it inherently converges to a single behavioral mean. Readers are referred to \cite{williams2017model} for details on MPPI.

\begin{figure*}[t]
  \centering
  \includegraphics[width=\linewidth,trim=0cm 1.5cm 0cm 0.5cm,clip]{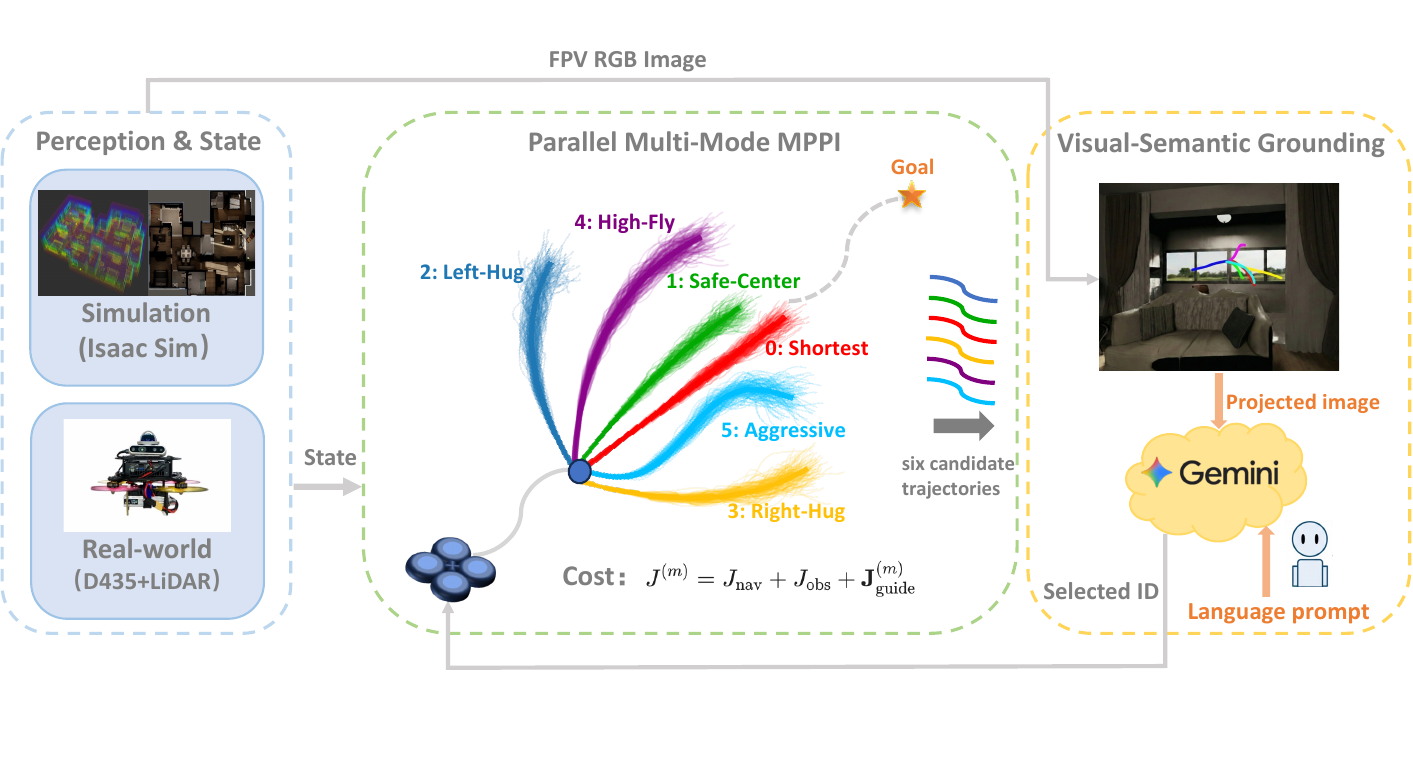}
  \vspace{-0.5cm}
  \caption{\textbf{System Architecture of the Language-Conditioned VLM-MPPI Framework.}
    The pipeline integrates high-frequency dynamic planning with semantic visual reasoning.
    \textbf{(Left) Perception \& State:} State estimation is obtained from Isaac Sim (simulation) or LiDAR-inertial odometry (real-world).
    \textbf{(Middle) Parallel Multi-Mode MPPI:} This module simultaneously optimizes distinct trajectory candidates. By incorporating mode-specific guiding terms $J_{\text{guide}}^{(m)}$ into the cost function, we induce distinct behavioral modes that converge to unique spatial means, rather than relying on random stochastic variations.
    \textbf{(Right) Visual-Semantic Grounding:} The 3D candidates are projected onto the FPV RGB image to reformulate navigation as a visual selection task. The \textbf{VLM (Gemini 2.5)} interprets the user's natural language prompt and selects the corresponding mode index, which is then tracked by a PID controller.}
    \vspace{-14pt}
  \label{fig:system_overview}
\end{figure*}

\section{Methodology}
\subsection{System Overview}
\label{sec:system_overview}
As illustrated in Fig.~\ref{fig:system_overview}, our system forms a closed-loop hierarchy that maps natural-language commands to dynamically feasible UAV motion. In this work, we define a `Mode' as a distinct distribution of trajectories converging to a unique behavioral mean (e.g., adhering to a wall or maintaining high altitude). These modes are induced by parallel MPPI optimization threads, each governed by a specifically designed cost function $J^{(k)}$. Accordingly, at each planning step, trajectory candidates for six such behavioral modes are generated, with each mode consisting of 128 trajectories sampled from its specific distribution. We assume that these 6 modes are sufficient to span the major flight behaviors required for UAV navigation. For each mode, an optimal trajectory is obtained according to Eqs.\ref{eq:sec} - \ref{eq:forth}, conditioned on the target and the current environment, which is guaranteed to be collision-free by construction. This results in 6 feasible trajectory candidates at the current time step. These candidates are then projected onto the UAV’s first-person-view (FPV) image and provided to the Vision–Language Model (VLM), which selects the optimal trajectory based on the target and semantic instructions. The selected trajectory is subsequently executed by the UAV. To accommodate the relatively slow VLM while maintaining flight stability, we query the VLM on trajectories generated from a predicted future state along the currently executed MPPI trajectory, using an estimate of the VLM latency, while the MPPI planner and low-level controller run continuously at high rate on the last selected mode.

\subsection{Parallel Behavior-Aware MPPI}
\label{sec:parallel_mppi}

To bridge the gap between continuous control and discrete semantic instructions, we employ a parallelized ensemble of $K=6$ MPPI planners. While standard MPPI typically converges to a single global minimum, our parallel architecture maintains independent sampling distributions to capture distinct homotopy classes and behavioral styles. Specifically, we design six guiding modes to cover the full spectrum of spatial and dynamic maneuvers: Shortest and Safe-Center modes for baseline efficiency and safety; Left-Hug and Right-Hug modes for topological distinctness; High-Fly mode for vertical obstacle avoidance and Aggressive for higher speed flight. Each mode $k$ optimizes a specific cost function $J^{(k)}$ tailored to a distinct behavioral mode. The total cost is defined as:
\begin{equation}
    J^{(k)}(\mathbf{x}_t, \mathbf{u}_t) = J_{nav}(\mathbf{x}_t) + J_{coll}(\mathbf{x}_t) + \lambda \cdot J_{guide}^{(k)}(\mathbf{x}_t, \mathbf{u}_t)
\end{equation}
where $J_{nav}$ and $J_{coll}$ enforce common navigation goals and hard safety constraints based on the distance to the nearest obstacle. The coefficient $\lambda$ weights the behavioral adherence against the primary navigation objective. To induce behavioral diversity, the guiding term $J_{guide}^{(k)}$ for each mode is designed as follows:

\subsubsection{Shortest}
This mode prioritizes path efficiency. We minimize lateral deviation while rewarding forward velocity along the longitudinal direction $\mathbf{n}_{long}$:
\begin{equation}
    J_{guide}^{short} = w_{lat} ||\mathbf{e}_{lat}||^2 - \alpha_{fwd} (\mathbf{v}_t^\top \mathbf{n}_{long})
\end{equation}
where $\mathbf{e}_{lat}$ is the lateral position error vector, and $w_{lat}, \alpha_{fwd}$ are the lateral penalty weight and velocity gain, respectively.

\subsubsection{Safe-Center}
This mode maximizes safety by centering the drone in free space, leveraging the Euclidean Signed Distance Field (ESDF) for continuous collision awareness. We introduce a clearance reward based on the distance to the nearest obstacle $d(\mathbf{x}_t)$, limited by a saturation threshold $d_{max}$:
\begin{equation}
    J_{guide}^{safe} = w_{cen} ||\mathbf{e}_{lat}||^2 - w_{clr} \cdot \min(d(\mathbf{x}_t), d_{max})
\end{equation}
Here, $d(\mathbf{x}_t)$ denotes the signed distance queried from the ESDF map. The term $w_{cen}$ penalizes off-center drift, while $w_{clr}$ weights the clearance reward to actively push the trajectory away from obstacle boundaries.

\subsubsection{Left-Hug and Right-Hug}
These modes enforce wall-following by penalizing deviations from a desired lateral clearance $\delta_{ref}$. Let $e_{lat}$ be the scalar lateral deviation. The cost is:
\begin{equation}
    J_{guide}^{hug} = w_{trk} (e_{lat} - s_k \delta_{ref})^2 + w_{barr} \cdot \max(0, -s_k e_{lat})^2
\end{equation}
where $s_k = +1$ for \textit{Left-Hug} and $s_k = -1$ for \textit{Right-Hug}. The $\max(\cdot)$ function acts as a one-sided virtual barrier, penalizing the drone only if it crosses to the undesired side (opposite to the target direction $s_k$). $w_{trk}$ and $w_{barr}$ weight the tracking error and barrier violation, respectively.

\subsubsection{High-Fly}
To enable vertical obstacle avoidance without hovering, we couple altitude tracking with a time-varying velocity incentive:
\begin{equation}
    J_{guide}^{high} = w_z (z_t - z_{ref}(t))^2 - \alpha(t) v_{long} + \eta(t) ||\mathbf{v}_t||
\end{equation}
Here, $w_z$ is the altitude tracking weight, and $z_{ref}(t)$ interpolates to the desired clearance altitude. The longitudinal velocity $v_{long} = \mathbf{v}_t^\top \mathbf{n}_{long}$ is rewarded by a decaying weight $\alpha(t)$, while the damping term $\eta(t)$ grows over the horizon to ensure the drone stops precisely at the goal.

\subsubsection{Aggressive}
This mode encourages time-optimality by rewarding velocity along the geodesic direction $\mathbf{g}$ while relaxing safety margins:
\begin{equation}
    J_{guide}^{aggr} = -\beta \frac{\mathbf{v}_t^\top \mathbf{g}}{||\mathbf{g}||} + w_{soft} \left( \epsilon_{loose} - d(\mathbf{x}_t) \right)_+^2
\end{equation}
We use a reduced safety margin $\epsilon_{loose} < \epsilon_{nom}$ (where $\epsilon_{nom}$ denotes the standard safety distance) and a soft barrier function (using the $(\cdot)_+$ notation) to allow traversing narrow gaps at higher speeds, scaled by velocity gain $\beta$ and collision weight $w_{soft}$.

\subsubsection{Distribution Biasing}
Finally, we inject a \textit{time-decaying deterministic bias} into the sampling distribution:
\begin{equation}
    \mathbf{u}_t^{(k)} \sim \mathcal{N}(\mathbf{u}_{nom} + \gamma^t \cdot \boldsymbol{\mu}_{bias}^{(k)}, \Sigma)
\end{equation}
where $\gamma \in (0, 1)$ is a decay factor and $\Sigma$ is the exploration covariance. This bias $\boldsymbol{\mu}_{bias}^{(k)}$ guides the initial exploration into the desired homotopy class.

\subsection{Visual--Semantic Grounding}
\label{sec:visual_grounding}

To bridge the gap between the controller's metric search space and the VLM's semantic reasoning, we project the 3D MPPI candidate trajectories onto the UAV's egocentric image plane. This reformulation transforms a continuous control problem into a discrete visual selection task, leveraging the VLM's zero-shot spatial understanding capabilities.

\subsubsection{3D-to-2D Projection}
Let $\mathcal{T} = \{\tau^{(0)}, \dots, \tau^{(K-1)}\}$ denote the set of candidate trajectories, where $K=6$ corresponds to the defined navigation modes. Each trajectory consists of a sequence of discrete waypoints $\mathbf{p}_W \in \mathbb{R}^3$ in the global inertial frame. We map these points to the image plane via a chain of rigid-body transformations.
First, given the UAV's estimated body pose $(\mathbf{R}_{WB}, \mathbf{t}_{WB})$ in the world frame, a waypoint is transformed into the body frame as:
\begin{equation}
    \mathbf{p}_B = \mathbf{R}_{WB}^\top \bigl(\mathbf{p}_W - \mathbf{t}_{WB}\bigr)
\end{equation}
where $\mathbf{t}_{WB}$ denotes the position of the body origin in the world frame. Subsequently, we apply the extrinsic calibration $(\mathbf{R}_{CB}, \mathbf{t}_{BC})$ to transform the point into the camera optical frame:
\begin{equation}
    \mathbf{p}_C = \mathbf{R}_{CB} \bigl(\mathbf{p}_B - \mathbf{t}_{BC}\bigr) = [x_c, y_c, z_c]^\top
\end{equation}
where $\mathbf{t}_{BC}$ represents the camera's position relative to the body center. Finally, assuming a standard pinhole camera model with the optical axis aligned with the $z_c$-axis, the pixel coordinates $(u, v)$ are obtained via:
\begin{equation}
    u = f_x \frac{x_c}{z_c} + c_x,
    \qquad
    v = f_y \frac{y_c}{z_c} + c_y
\end{equation}
where $(f_x, f_y)$ and $(c_x, c_y)$ are the focal lengths and principal points, respectively.
To ensure visual consistency, a waypoint is valid only if $z_c > \delta_{\text{near}}$ (with $\delta_{\text{near}} = 0.05\,\text{m}$) and the projection lies within the image bounds.

\subsubsection{Visual Prompt Construction}
\begin{figure}[h]
    \centering
    \includegraphics[width=0.95\columnwidth]{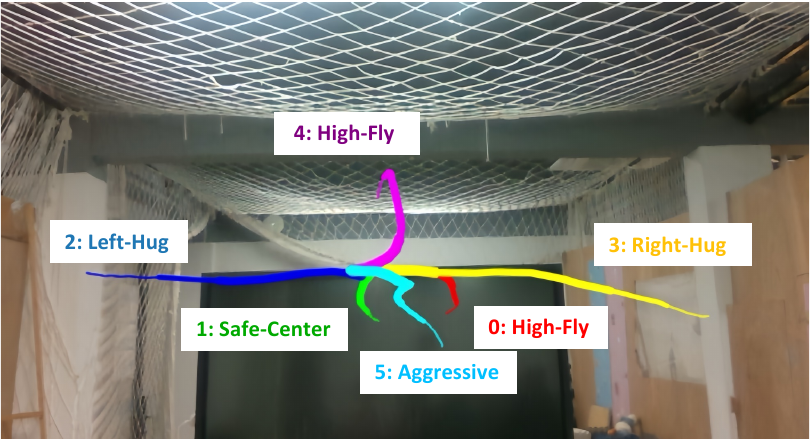}
    \caption{\textbf{Trajectory Projection for Visual Grounding.}  We project the 3D MPPI trajectory candidates onto the onboard FPV camera stream. The color-coded polylines serve as distinct visual prompts. }

    \label{fig:projection}
    \vspace{-10pt}
\end{figure}

The projected trajectories are rendered as color-coded polylines overlaid on the raw RGB image, as illustrated in Fig. \ref{fig:projection}. To anchor the VLM's semantic reasoning, we establish a fixed color--behavior mapping. By omitting textual labels on the image, we maximize visual clarity and minimize occlusions of environmental features. Instead, the textual prompt provides descriptions of each navigation mode (0--5) and their corresponding colors. This design reformulates the navigation task into a multiple-choice selection problem, where the VLM identifies the most appropriate behavioral mode that aligns the visualized geometry with the user's high-level instruction.

\section{Experiments}
\label{sec:experiments}

\subsection{Experimental Setup}
\label{sec:experimental-setup}

We evaluate the proposed framework in high-fidelity simulation using NVIDIA Isaac Sim integrated with a ROS~1 communication stack. The UAV is modeled as a 6-DoF quadrotor controlled by a position--yaw PID controller running at $50\,\mathrm{Hz}$, which tracks reference poses generated by the planner and outputs body-frame velocity commands. Static indoor environments are pre-processed into signed distance fields (SDFs) for real-time collision checking. An onboard monocular camera provides $640\times480$ first-person-view (FPV) images. A dedicated projection node maps candidate trajectories from the multi-mode MPPI planner onto the image plane, providing spatial grounding for the high-level vision--language decision maker.

The hierarchical control pipeline operates as follows. The parallel MPPI planner generates six behaviorally distinct trajectories at $20\,\mathrm{Hz}$, while the low-level PID controller ensures stable tracking. The planner samples $N=128$ control sequences over a horizon of $H=32$ steps. The resulting modes---\textit{Shortest}, \textit{Safe-Center}, \textit{Left-Hug}, \textit{Right-Hug}, \textit{High-Fly}, and \textit{Aggressive}---are rendered as an FPV overlay and passed, together with the user command, to a pre-trained Gemini~2.5 Flash Vision--Language Model, which selects the mode index to execute. In all experiments, the MPPI planner (implemented in PyTorch) and control stack run entirely on the CPU in real time; a single NVIDIA GeForce RTX 5090 GPU is used for Isaac Sim rendering.

\subsection{Multi-Modal MPPI Behavior Characterization}
\label{sec:mppi_modes}

Before integrating the VLM, we first characterize the behavioral modes induced purely by the Parallel Behavior-Aware MPPI. In a fixed indoor environment with a common start and goal, we execute 30 autonomous flight trials, running each mode independently for 5 runs under the same geometric goal. For each trajectory, we record the path length, average speed, average lateral offset from the nominal longitudinal axis, and the spatial root-mean-square error (RMSE) within the mode, which measures the dispersion of trajectories in configuration space.

\begin{table}[h]
    \centering
    \caption{Quantitative statistics for each MPPI behavioral mode over 5 trials (mean $\pm$ standard deviation).}
    \label{tab:mppi_results_final}
    \addtolength{\tabcolsep}{-2pt}
    \resizebox{\columnwidth}{!}{
        \begin{tabular}{lcccc}
            \toprule
            \textbf{Behavioral} & \textbf{Path Length} & \textbf{Avg. Velocity} & \textbf{Avg. Lat. Offset} & \textbf{Intra-mode} \\
            \textbf{Mode}       & [m]                  & [m/s]                  & [m]                       & \textbf{RMSE [m]}   \\
            \midrule
            Shortest            & $8.30 \pm 0.08$      & $1.06 \pm 0.02$        & $0.15 \pm 0.03$           & \textbf{0.091}      \\
            Safe Center         & $8.36 \pm 0.07$      & $0.97 \pm 0.03$        & $0.19 \pm 0.01$           & \textbf{0.090}      \\
            Left Hug            & $8.83 \pm 0.34$      & $1.03 \pm 0.05$        & $0.46 \pm 0.03$           & 0.131               \\
            Right Hug           & $8.91 \pm 0.12$      & $1.01 \pm 0.03$        & $0.88 \pm 0.03$           & 0.158               \\
            High Fly            & $12.49 \pm 1.22$     & $0.95 \pm 0.02$        & $0.33 \pm 0.03$           & 0.215               \\
            \textbf{Aggressive} & \textbf{8.39} $\pm 0.23$ & \textbf{1.27} $\pm 0.02$ & $0.17 \pm 0.04$       & 0.215               \\
            \bottomrule
        \end{tabular}
    }
    \vspace{-1em}
\end{table}

\begin{figure}[H]
    \centering

    \includegraphics[width=\linewidth,trim=5cm 2cm 5cm 1.8cm,clip]{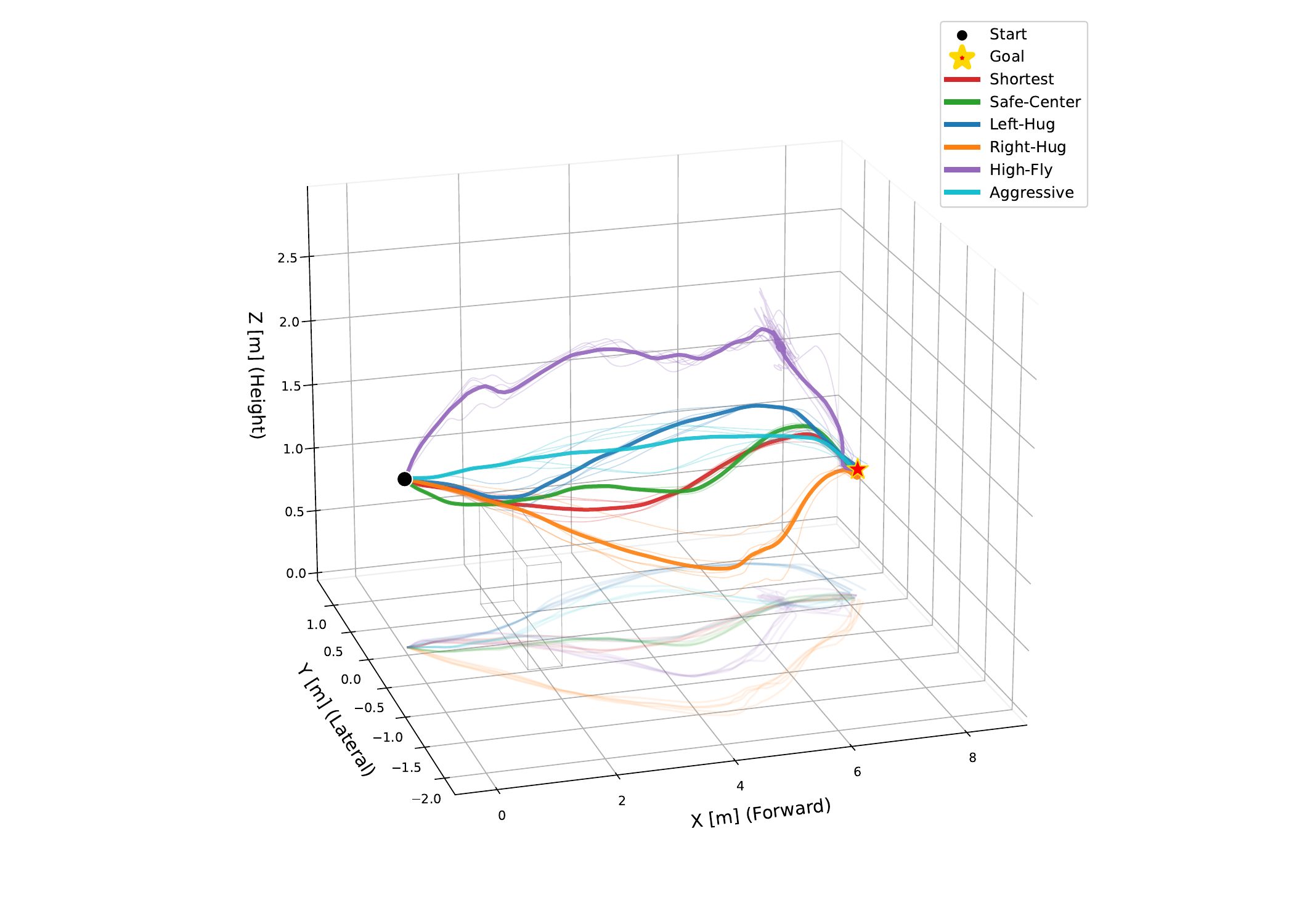}
    \caption{\textbf{Behavioral Diversity:} 3D visualization of 30 autonomous trajectories generated by the proposed multi-modal MPPI. Ground projections ($z=0$) demonstrate clear spatial segregation among the trajectory clusters, confirming that the mode-specific guiding costs successfully induce physically distinct behavioral distributions rather than mere stochastic variations.}
    \label{fig:mppi_3d_topo}
    \vspace{-10pt}
\end{figure}

The resulting spatial distributions are visualized in Fig.~\ref{fig:mppi_3d_topo}, while Table~\ref{tab:mppi_results_final} summarizes the quantitative statistics across all modes. The six MPPI modes realize clear and physically consistent motion patterns that match their intended semantics. The Aggressive mode achieves the highest mean speed, $1.27 \pm 0.02\,\mathrm{m/s}$, corresponding to an increase of about $30\%$ over the conservative Safe-Center baseline ($0.97 \pm 0.03\,\mathrm{m/s}$), while maintaining a path length comparable to the shortest route. The Left-Hug and Right-Hug modes exhibit well-separated average lateral offsets of $0.46 \pm 0.03\,\mathrm{m}$ and $0.88 \pm 0.03\,\mathrm{m}$, respectively, confirming that the guiding costs successfully encode distinct wall-following behaviors. At the same time, the intra-mode RMSE values remain small: the Shortest and Safe-Center modes yield mean spatial RMSEs of approximately $0.09\,\mathrm{m}$, comparable to the UAV's physical dimensions, and even the more aggressive modes stay below $0.22\,\mathrm{m}$. All modes achieve collision-free navigation in all 30 rollouts, indicating that the induced behavioral diversity does not compromise safety.

\subsection{Quantitative Evaluation of Language-Conditioned Navigation}
\label{sec:full-pipeline-quant}

Building on the MPPI-only analysis, we now quantify the behavior of the full language-conditioned VLM--MPPI system. The dynamics model, MPPI hyperparameters, and indoor apartment layout remain unchanged; the only difference is that the behavioral mode is selected online by the VLM in response to natural-language commands (e.g., ``fly to the goal efficiently'', ``fly along the right side'', ``fly higher to avoid obstacles''). For each of the six modes, we execute five trials with fixed start and goal configurations, yielding $30$ language-conditioned flights in total.

\begin{table*}[t]
\centering
\caption{\textbf{Quantitative Evaluation across Behavioral Modes.} }
\label{tab:quant_results}
\resizebox{\textwidth}{!}{\begin{tabular}{l c c c c c c}
\toprule
\textbf{Mode} & \textbf{Time (s)} & \textbf{Path Len. (m)} & \textbf{Latency (s)} & \textbf{Safety (m)} & \textbf{Lat. Offset (m)} & \textbf{RMSE (m)} \\
\midrule
Shortest     & $9.47 \pm 0.35$ & $8.41 \pm 0.29$ & $2.64 \pm 0.25$ & $0.03 \pm 0.00$ & $-0.13 \pm 0.02$ & $0.115$ \\
Safe Center  & $10.75 \pm 0.46$ & $8.34 \pm 0.14$ & $2.86 \pm 0.04$ & $0.04 \pm 0.00$ & $-0.19 \pm 0.01$ & $\mathbf{0.058}$ \\
Left Hug     & $11.18 \pm 1.14$ & $8.85 \pm 0.19$ & $2.76 \pm 0.16$ & $0.03 \pm 0.00$ & $0.24 \pm 0.02$  & $0.110$ \\
Right Hug    & $11.17 \pm 1.15$ & $8.94 \pm 0.11$ & $2.63 \pm 0.12$ & $0.03 \pm 0.00$ & $\mathbf{-0.64 \pm 0.06}$ & $0.111$ \\
High Fly     & $16.55 \pm 3.78$ & $10.64 \pm 1.24$ & $2.75 \pm 0.10$ & $0.02 \pm 0.00$ & $0.13 \pm 0.41$ & $0.641$ \\
Aggressive   & $\mathbf{7.90 \pm 0.24}$ & $8.26 \pm 0.07$ & $2.36 \pm 0.15$ & $0.03 \pm 0.00$ & $-0.15 \pm 0.06$ & $0.110$ \\
\bottomrule
\end{tabular}}
\end{table*}

\begin{figure*}[h]
    \centering
    \includegraphics[width=0.98\textwidth]{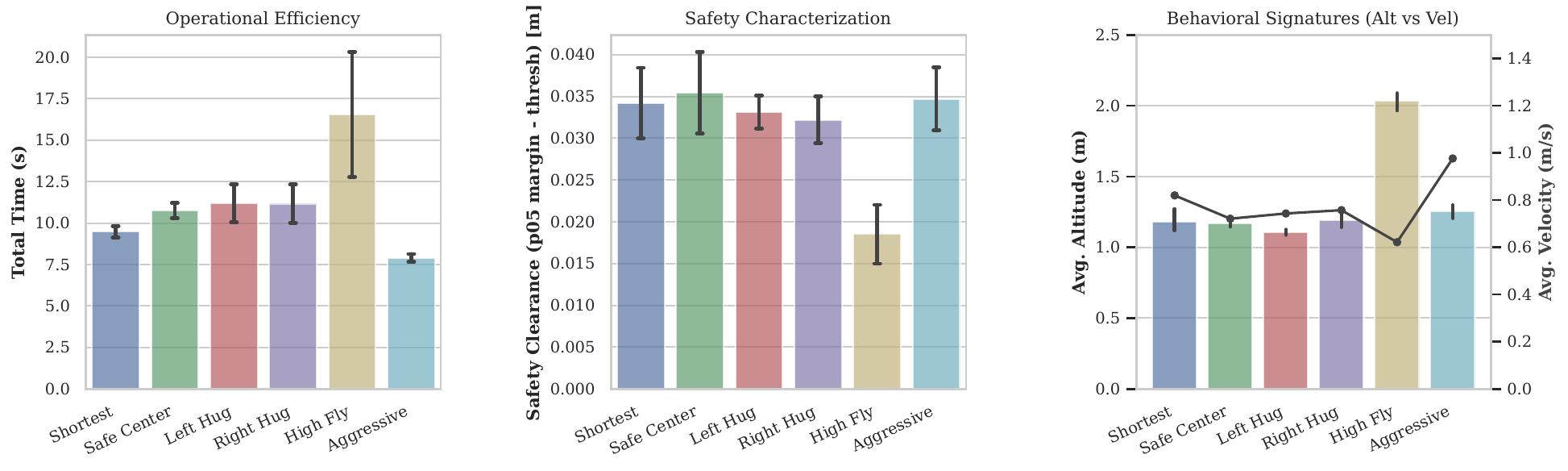}
    \caption{\textbf{Quantitative behavior of the full language-conditioned VLM--MPPI pipeline.}
    Data aggregates 5 trials per mode.
    \textbf{Left (Operational Efficiency):} Active flight duration (excluding start/end idle periods). The Aggressive mode achieves the shortest duration, validating its time-optimality.
    \textbf{Middle (Safety Characterization):} The $5^{th}$-percentile safety clearance relative to the hard collision threshold ($d_{th} = -0.12\,\mathrm{m}$). All modes maintain positive margins, demonstrating that behavioral diversity does not compromise safety.
    \textbf{Right (Behavioral Signatures):} Average altitude (bars, left axis) overlaid with average velocity (line, right axis). The High-Fly mode distinctly occupies the upper airspace ($\approx 1.9\,\mathrm{m}$) at lower speeds, whereas Aggressive maintains nominal altitude but maximizes cruise velocity.
    These trends confirm that VLM-selected modes produce distinct behaviors while the MPPI backend ensures safety.}
    \label{fig:vlm_dual_path_results}
    \vspace{-12pt}
\end{figure*}

\begin{figure}[H]
  \centering
  \includegraphics[width=\columnwidth]{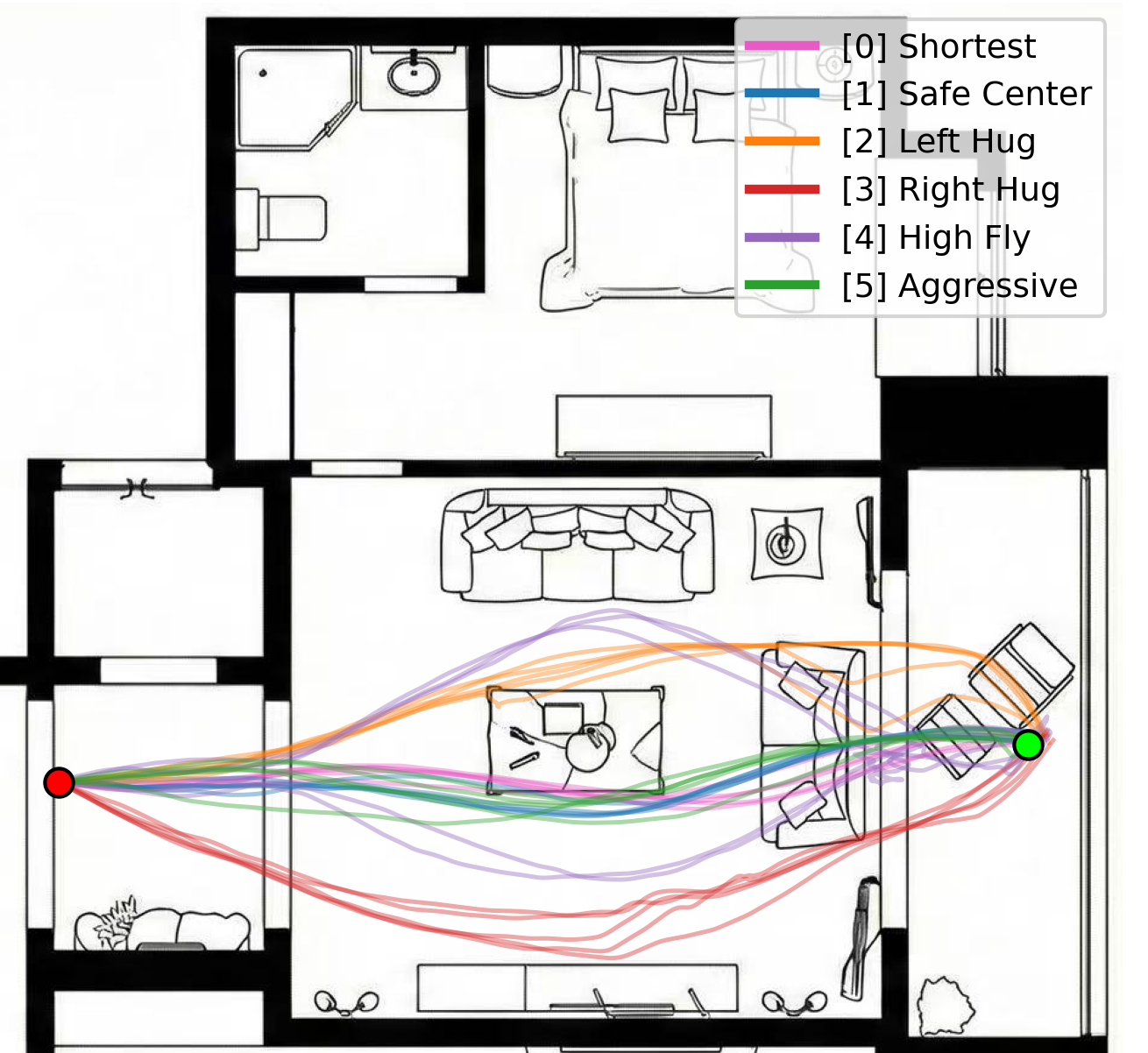}
  \caption{\textbf{Top–down view of behavioral diversity (5 trials per mode).}
    The system consistently generates distinct geometric styles, such as wall-hugging and aggressive flight.}
  \label{fig:topdown_lang}
  \vspace{-12pt}
\end{figure}

Figure~\ref{fig:topdown_lang} provides a complementary top–down visualization of all $30$ runs. For a fixed start and goal, the colored bundles form clearly separated geometric styles that match their intended semantics. The High-Fly mode is particularly illustrative: its trajectories split into two homotopy classes (passing left or right) while maintaining elevated altitude, showing that the prompt ``fly higher'' strictly constrains vertical behavior while leaving lateral route choice to the MPPI optimizer's local minima.

Figure~\ref{fig:vlm_dual_path_results} summarizes the aggregate performance metrics. The left panel confirms that the Aggressive mode is the fastest ($7.90 \pm 0.24$\,s), reducing active flight duration by approximately 17\% compared to the baseline Shortest mode ($9.47 \pm 0.35$\,s), whereas High-Fly is the slowest ($16.55 \pm 3.78$\,s) due to the deliberate vertical maneuvers required to reach the clearance altitude. As shown in the middle panel, the Safety Characterization plots the $5^{th}$-percentile safety margin relative to the collision threshold; a key finding is that all behavioral modes maintain positive clearance values ($0.02$--$0.04$\,m above the hard threshold). This quantitatively validates the robustness of our hierarchical framework: despite the VLM triggering diverse and potentially aggressive behaviors, the underlying MPPI controller consistently enforces safety constraints. Finally, the right panel highlights the decoupling of geometric and dynamic objectives: the High-Fly mode is characterized by its elevated altitude ($\approx 1.9$\,m) and moderate speed, whereas the \textit{Aggressive} mode maintains a standard altitude but achieves the highest average velocity ($\approx 1.05$\,m/s), significantly outpacing the conservative Safe-Center mode ($\approx 0.78$\,m/s).

Table~\ref{tab:quant_results} provides further numerical evidence of Intent Alignment and system performance. The recorded latency ($\approx 2.4\text{--}2.9$\,s) primarily reflects the VLM inference and network transmission time; notably, the parallel MPPI planner ensures flight safety during these asynchronous updates. The Lat. Offset column reveals a significant sign difference between Left-Hug ($0.24 \pm 0.02$\,m) and Right-Hug ($-0.64 \pm 0.06$\,m), confirming that the VLM's selection successfully guides the MPPI optimizer into topologically distinct homotopy classes. Furthermore, the low intra-mode RMSE values ($\approx 0.06$--$0.12$\,m) for most modes indicate high repeatability. The larger RMSE for High-Fly ($0.641$\,m) is expected, reflecting the greater kinematic freedom available in the collision-free space above the obstacles.

\begin{figure*}[t]
  \centering
  \includegraphics[width=0.98\textwidth]{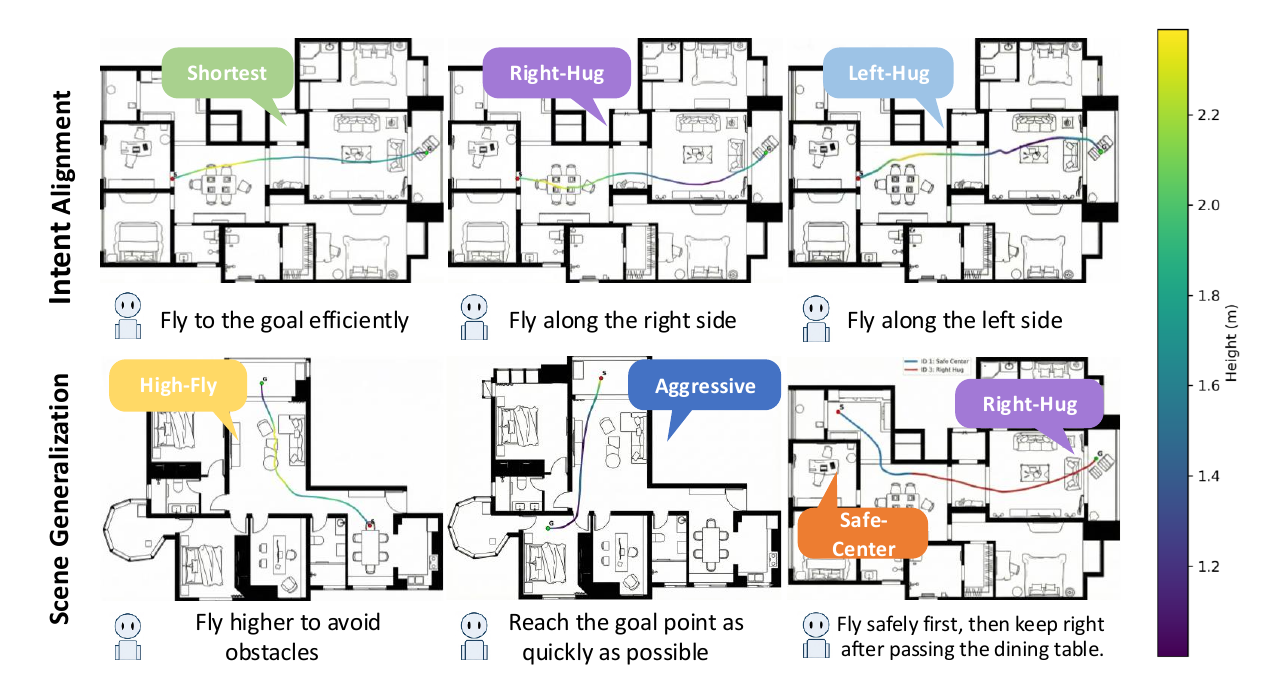}
  \caption{\textbf{Qualitative results of language-conditioned UAV navigation.}
  The figure illustrates how the proposed VLM--MPPI framework bridges high-level linguistic instructions with physical motion execution.
  \textbf{Row 1 (Intent Alignment):} topological path bifurcations (Shortest, Right-Hug, Left-Hug) in a fixed scenario under varied prompts.
  \textbf{Row 2 (Scene Generalization):} zero-shot adaptation across different indoor layouts, including high-altitude flight and aggressive maneuvering.
  The rightmost panel highlights a dynamic mode switch: the agent initially maintains a Safe-Center policy and then transitions to right-wall hugging upon receiving a mid-flight corrective command.}
  \label{fig:qualitative_results}
  \vspace{-1.5em}
\end{figure*}

Complementing these metrics, Fig.~\ref{fig:qualitative_results} visually demonstrates the system's Intent Alignment and Scene Generalization. As shown in the top row, changing the prompt effectively alters the homotopy class (e.g., from Mode 0 to Mode 3) without changing the geometric goal. The framework also supports asynchronous mid-flight switching (Fig.~\ref{fig:qualitative_results}, bottom right), validating the system's responsiveness to evolving user intent. Overall, these results demonstrate that the VLM--MPPI system preserves the safety and consistency of the underlying controller while robustly aligning physical behaviors with high-level semantic instructions.

\subsection{Real-World Experiments}
\label{sec:real_world_exp}

To validate the Sim-to-Real transferability of our behavioral shaping approach, we conducted flight tests on a custom-built quadrotor platform. The onboard computation is powered by an Intel NUC 13 (running Ubuntu 20.04 with ROS Noetic). State estimation is performed using the FAST-LIO2 algorithm~\cite{xu2022fast}, processing point clouds from a Livox Mid-360 LiDAR. For visual-semantic grounding, we utilize a RealSense D435 camera to employ its RGB image for VLM inference, verifying that our high-level decision-making relies on semantic reasoning rather than dense depth maps. Low-level attitude control is executed by a Betaflight flight controller via a serial interface. The experiments were performed in a $7\,\text{m} \times 4\,\text{m} \times 3\,\text{m}$ indoor environment containing everyday obstacles such as chairs, a TV, and boxes.

\begin{figure}[H]
    \vspace{-1em}
    \centering

    \includegraphics[width=\linewidth,trim=0 2cm 1cm 0cm,clip]{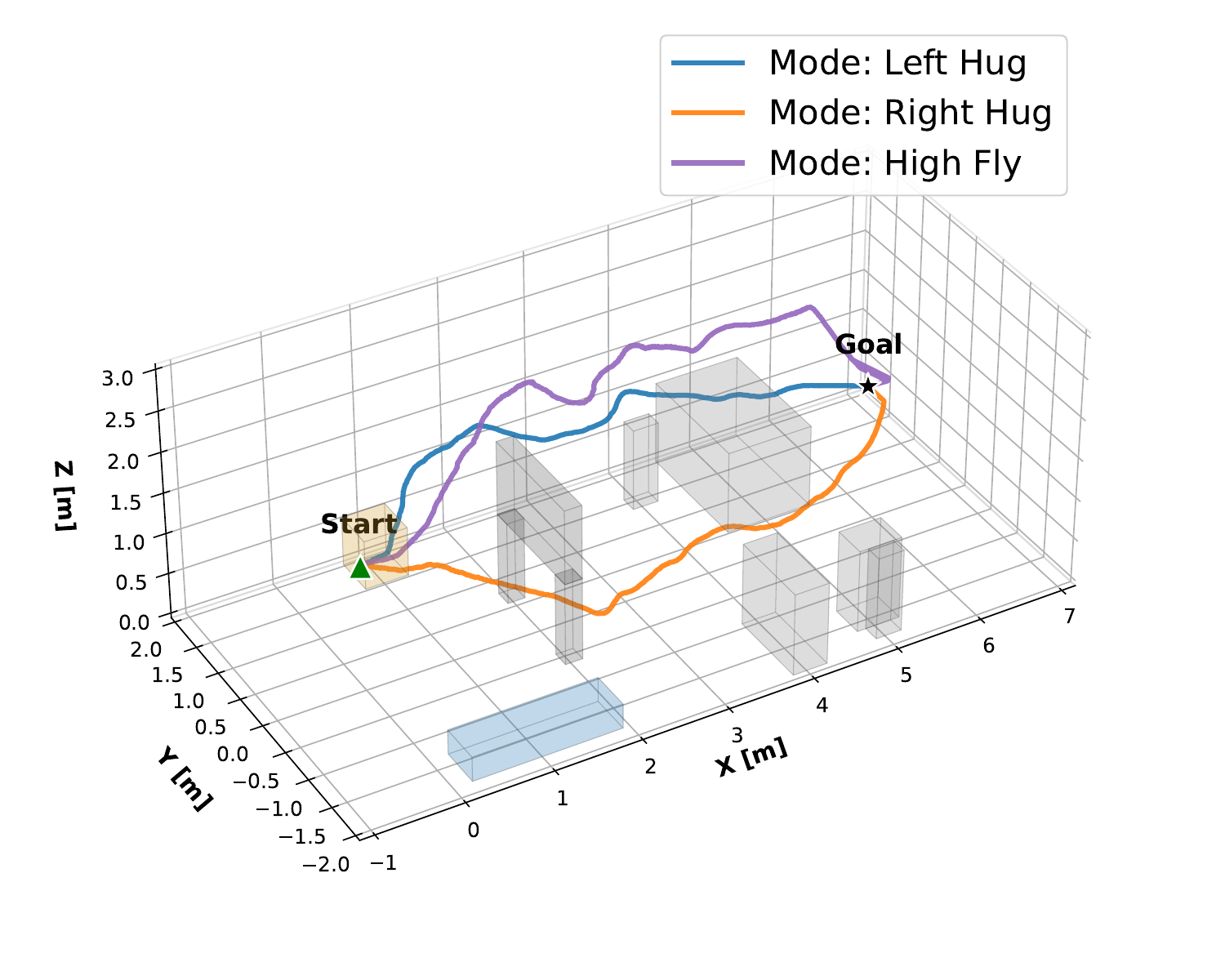}
    \caption{\textbf{Real-World Trajectory Execution.} Overlay of flown paths collected from the physical quadrotor. The system demonstrates robust behavioral segregation: Left-Hug (Blue) and Right-Hug (Yellow) maintain safe lateral clearances, while the High-Fly mode (Purple) effectively utilizes the vertical space ($z > 1.5\,\mathrm{m}$) to overfly obstacles, verifying the feasibility of 3D semantic navigation in physical environments.}
    \label{fig:real_world_traj}
    \vspace{-12pt}
\end{figure}

We evaluated three topologically distinct modes: Left-Hug, Right-Hug, and High-Fly, to verify that the distinct cost landscapes designed in simulation remain effective under real-world sensing noise and actuation disturbances.
Fig.~\ref{fig:real_world_traj} visualizes the composite trajectories from multiple runs.

\begin{table}[h]
    \centering
    \caption{\textbf{Real-World Results.} }
    \label{tab:real_world_stats}
    \scriptsize
    \setlength{\tabcolsep}{8pt}
    \renewcommand{\arraystretch}{1}
    \begin{tabular}{lccc}
        \toprule
        \textbf{Mode} & \textbf{Speed (m/s)} & \textbf{Height (m)} & \textbf{Offset (m)} \\
        \midrule
        Left Hug  & $0.32$ & $1.58$ & $\mathbf{-0.75}$ \\
        Right Hug & $0.33$ & $1.54$ & $\mathbf{+0.39}$ \\
        High Fly  & $0.39$ & $\mathbf{2.02}$ & $-0.28$ \\
        \bottomrule
    \end{tabular}
    \vspace{-2em}
\end{table}

Table~\ref{tab:real_world_stats} summarizes the results.
The system achieved a 100\% collision-free rate across all 30 trials, validating the safety of the proposed planner.
Although the average speed ($\approx 0.35\,\mathrm{m/s}$) was kept conservative for safety in the confined physical space, the behavioral distinctions remained prominent. The High-Fly mode reached an average altitude of $\mathbf{2.02\,\mathrm{m}}$, successfully clearing the vertical space above obstacles. The MPPI planner demonstrated strong robustness against disturbances through its high-frequency receding horizon replanning, ensuring collision-free flight in all trials. The Left-Hug and Right-Hug modes exhibited distinct average lateral offsets of $-0.75\,\mathrm{m}$ and $+0.39\,\mathrm{m}$ respectively.

These results confirm that our cost function design is robust enough to enforce high-level behavioral constraints in the physical world without requiring extensive parameter retuning.

\section{CONCLUSION}
In this work, we presented a hierarchical framework for language-conditioned aerial navigation that effectively bridges the gap between high-level semantic intent and low-level dynamic control. The core novelty of our approach lies in the generation of behaviorally diverse navigation modes: rather than relying on stochastic exploration, we design mode-specific guiding costs that induce distinct behavioral distributions, each converging to a unique mean. This allows the system to offer a discrete menu of physically feasible modes to the VLM for selection. Extensive experiments in both Isaac Sim and on a real-world quadrotor platform demonstrate that this paradigm achieves robust intent alignment in cluttered indoor environments. The system successfully grounds natural language commands into complex topological behaviors and zero-shot generalizations without requiring task-specific retraining. Future work will focus on extending the parallel MPPI formulation to handle dynamic obstacles (e.g., moving pedestrians) and integrating an onboard VLM for fully map-less navigation in unstructured environments.

\addtolength{\textheight}{-12cm}

\bibliographystyle{ieeetr}
\bibliography{refs}

@article{driess2023palm,
  title={Palm-e: An embodied multimodal language model},
  author={Driess, Danny and Xia, Fei and Sajjadi, Mehdi SM and Lynch, Corey and Chowdhery, Aakanksha and Ichter, Brian and Wahid, Ayzaan and Tompson, Jonathan and Vuong, Quan and Yu, Tianhe and others},
  journal={arXiv preprint arXiv:2303.03378},
  year={2023}
}

@inproceedings{gu2022vision,
  title={Vision-and-language navigation: A survey of tasks, methods, and future directions},
  author={Gu, Jing and Stefani, Eliana and Wu, Qi and Thomason, Jesse and Wang, Xin},
  booktitle={Proceedings of the 60th Annual Meeting of the Association for Computational Linguistics (Volume 1: Long Papers)},
  pages={7606--7623},
  year={2022}
}

@article{duan2022survey,
  title={A survey of embodied ai: From simulators to research tasks},
  author={Duan, Jiafei and Yu, Samson and Tan, Hui Li and Zhu, Hongyuan and Tan, Cheston},
  journal={IEEE Transactions on Emerging Topics in Computational Intelligence},
  volume={6},
  number={2},
  pages={230--244},
  year={2022},
  publisher={IEEE}
}

@inproceedings{mellinger2011minimum,
  title={Minimum snap trajectory generation and control for quadrotors},
  author={Mellinger, Daniel and Kumar, Vijay},
  booktitle={2011 IEEE international conference on robotics and automation},
  pages={2520--2525},
  year={2011},
  organization={Ieee}
}

@article{zhou2020ego,
  title={Ego-planner: An esdf-free gradient-based local planner for quadrotors},
  author={Zhou, Xin and Wang, Zhepei and Ye, Hongkai and Xu, Chao and Gao, Fei},
  journal={IEEE Robotics and Automation Letters},
  volume={6},
  number={2},
  pages={478--485},
  year={2020},
  publisher={IEEE}
}

@article{zhao2025rethinking,
  title={Rethinking Reference Trajectories in Agile Drone Racing: A Unified Reference-Free Model-Based Controller via MPPI},
  author={Zhao, Fangguo and Guan, Xin and Li, Shuo},
  journal={arXiv preprint arXiv:2509.14726},
  year={2025}
}

@article{zhou2019robust,
  title={Robust and efficient quadrotor trajectory generation for fast autonomous flight},
  author={Zhou, Boyu and Gao, Fei and Wang, Luqi and Liu, Chuhao and Shen, Shaojie},
  journal={IEEE Robotics and Automation Letters},
  volume={4},
  number={4},
  pages={3529--3536},
  year={2019},
  publisher={IEEE}
}

@article{achiam2023gpt,
  title={Gpt-4 technical report},
  author={Achiam, Josh and Adler, Steven and Agarwal, Sandhini and Ahmad, Lama and Akkaya, Ilge and Aleman, Florencia Leoni and Almeida, Diogo and Altenschmidt, Janko and Altman, Sam and Anadkat, Shyamal and others},
  journal={arXiv preprint arXiv:2303.08774},
  year={2023}
}

@article{comanici2025gemini,
  title={Gemini 2.5: Pushing the frontier with advanced reasoning, multimodality, long context, and next generation agentic capabilities},
  author={Comanici, Gheorghe and Bieber, Eric and Schaekermann, Mike and Pasupat, Ice and Sachdeva, Noveen and Dhillon, Inderjit and Blistein, Marcel and Ram, Ori and Zhang, Dan and Rosen, Evan and others},
  journal={arXiv preprint arXiv:2507.06261},
  year={2025}
}

@inproceedings{Lm-nav,
  title={Lm-nav: Robotic navigation with large pre-trained models of language, vision, and action},
  author={Shah, Dhruv and Osi{\'n}ski, B{\l}a{\.z}ej and Levine, Sergey and others},
  booktitle={Conference on robot learning},
  pages={492--504},
  year={2023},
  organization={PMLR}
}

@article{Clip-nav,
  title={Clip-nav: Using clip for zero-shot vision-and-language navigation},
  author={Dorbala, Vishnu Sashank and Sigurdsson, Gunnar and Piramuthu, Robinson and Thomason, Jesse and Sukhatme, Gaurav S},
  journal={arXiv preprint arXiv:2211.16649},
  year={2022}
}

@inproceedings{Convoi,
  title={Convoi: Context-aware navigation using vision language models in outdoor and indoor environments},
  author={Sathyamoorthy, Adarsh Jagan and Weerakoon, Kasun and Elnoor, Mohamed and Zore, Anuj and Ichter, Brian and Xia, Fei and Tan, Jie and Yu, Wenhao and Manocha, Dinesh},
  booktitle={2024 IEEE/RSJ International Conference on Intelligent Robots and Systems (IROS)},
  pages={13837--13844},
  year={2024},
  organization={IEEE}
}

@article{language,
  title={Language as Cost: Proactive Hazard Mapping using VLM for Robot Navigation},
  author={Oh, Mintaek and Kim, Chan and Seo, Seung-Woo and Kim, Seong-Woo},
  journal={arXiv preprint arXiv:2508.03138},
  year={2025}
}

@article{garg2026online,
  title={Online Motion Planning for Connected Multi-Robot Systems using Vision Language Models as High-level Planners},
  author={Garg, Kunal and Nair, Devika Shaj Kumar and Zhang, Songyuan and Arkin, Jacob and Fan, Chuchu},
  journal={Proceedings of Machine Learning Research vol vvv},
  volume={1},
  pages={21},
  year={2026}
}

@article{UAV-VLPA,
  title={UAV-VLPA*: A Vision-Language-Path-Action System for Optimal Route Generation on a Large Scales},
  author={Sautenkov, Oleg and Akhmetkazy, Aibek and Yaqoot, Yasheerah and Mustafa, Muhammad Ahsan and Tadevosyan, Grik and Lykov, Artem and Tsetserukou, Dzmitry},
  journal={arXiv preprint arXiv:2503.02454},
  year={2025}
}

@article{ComposableNav,
  title={ComposableNav: Instruction-Following Navigation in Dynamic Environments via Composable Diffusion},
  author={Hu, Zichao and Tang, Chen and Munje, Michael J and Zhu, Yifeng and Liu, Alex and Liu, Shuijing and Warnell, Garrett and Stone, Peter and Biswas, Joydeep},
  journal={arXiv preprint arXiv:2509.17941},
  year={2025}
}

@article{visiopath,
  title={VisioPath: Vision-Language Enhanced Model Predictive Control for Safe Autonomous Navigation in Mixed Traffic},
  author={Wang, Shanting and Typaldos, Panagiotis and Li, Chenjun and Malikopoulos, Andreas A},
  journal={IEEE Open Journal of Control Systems},
  year={2025},
  publisher={IEEE}
}

@inproceedings{
        rana2023sayplan,
        title={SayPlan: Grounding Large Language Models using 3D Scene Graphs for Scalable Task Planning},
        author={Krishan Rana and Jesse Haviland and Sourav Garg and Jad Abou-Chakra and Ian Reid and Niko Suenderhauf},
        booktitle={7th Annual Conference on Robot Learning},
        year={2023},
        url={https://openreview.net/forum?id=wMpOMO0Ss7a}
      }

@inproceedings{chen2023open,
  title={Open-vocabulary queryable scene representations for real world planning},
  author={Chen, Boyuan and Xia, Fei and Ichter, Brian and Rao, Kanishka and Gopalakrishnan, Keerthana and Ryoo, Michael S and Stone, Austin and Kappler, Daniel},
  booktitle={2023 IEEE International Conference on Robotics and Automation (ICRA)},
  pages={11509--11522},
  year={2023},
  organization={IEEE}
}

@article{huang2023voxposer,
      title={VoxPoser: Composable 3D Value Maps for Robotic Manipulation with Language Models},
      author={Huang, Wenlong and Wang, Chen and Zhang, Ruohan and Li, Yunzhu and Wu, Jiajun and Fei-Fei, Li},
      journal={arXiv preprint arXiv:2307.05973},
      year={2023}
    }

@article{williams2017model,
  title={Model predictive path integral control: From theory to parallel computation},
  author={Williams, Grady and Aldrich, Andrew and Theodorou, Evangelos A},
  journal={Journal of Guidance, Control, and Dynamics},
  volume={40},
  number={2},
  pages={344--357},
  year={2017},
  publisher={American Institute of Aeronautics and Astronautics}
}

@article{AERO-MPPI,
  title={AERO-MPPI: Anchor-Guided Ensemble Trajectory Optimization for Agile Mapless Drone Navigation},
  author={Chen, Xin and Huang, Rui and Tang, Longbin and Zhao, Lin},
  journal={arXiv preprint arXiv:2509.17340},
  year={2025}
}

@article{Pa-mppi,
  title={Pa-mppi: Perception-aware model predictive path integral control for quadrotor navigation in unknown environments},
  author={Zhai, Yifan and Reiter, Rudolf and Scaramuzza, Davide},
  journal={arXiv preprint arXiv:2509.14978},
  year={2025}
}

@article{xu2022fast,
  title={Fast-lio2: Fast direct lidar-inertial odometry},
  author={Xu, Wei and Cai, Yixi and He, Dongjiao and Lin, Jiarong and Zhang, Fu},
  journal={IEEE Transactions on Robotics},
  volume={38},
  number={4},
  pages={2053--2073},
  year={2022},
  publisher={IEEE}
}

\end{document}